\documentclass[a4paper,fleqn]{cas-dc}

\usepackage[authoryear,longnamesfirst]{natbib}

\def\tsc#1{\csdef{#1}{\textsc{\lowercase{#1}}\xspace}}
\tsc{WGM}
\tsc{QE}
\tsc{EP}
\tsc{PMS}
\tsc{BEC}
\tsc{DE}

\begin{document}
\let\WriteBookmarks\relax
\def\floatpagepagefraction{1}
\def\textpagefraction{.001}

\shorttitle{Less Is More}

\title{Less Is More: An Explainable AI Framework for Lightweight Malaria Classification}

\author[1]{Md Abdullah Al Kafi}
\ead{kafi.cse@diu.edu.bd}

\author[1]{Raka Moni}
\ead{rakamoni509@diu.edu.bd}

\author[2]{Sumit Kumar Banshal\cormark[1]}
\ead{sumitbanshal06@gmail.com}

\address[1]{Daffodil International University, Dhaka, Bangladesh}
\address[2]{Alliance University, Bengaluru, India}

\cortext[1]{Corresponding author. Email: sumitbanshal06@gmail.com}

\begin{abstract}
\textbf{Background and Objective:}
Deep learning models have high computational needs and lack interpretability, but are often the first choice for medical image classification tasks. Here, this study addresses whether complex neural networks are essential for the simple binary classification task of malaria. We introduce the Extracted Morphological Feature Engineered (EMFE) pipeline, a transparent, reproducible, and low-compute machine learning approach tailored explicitly for simple cell morphology, which was designed to achieve deep learning performance levels on a simple CPU-only setup with the practical aim of real-world deployment.
\textbf{Methods:}
The study used the NIH Malaria Cell Images dataset, with two features extracted from each cell image: the number of non-background pixels and the number of holes within the cell. Logistic Regression and Random Forest were compared against ResNet18, DenseNet121, MobileNetV2, and EfficientNet across accuracy, model size, and CPU inference time. An ensemble model was then created by combining Logistic Regression and Random Forests to achieve higher accuracy while retaining efficiency.
\textbf{Results:}
The single-variable Logistic Regression model achieved a test accuracy of 94.80\% with a file size of just 1.2 kB and negligible inference latency (~2.3 ms). The two-stage ensemble further improved accuracy to 97.15\%. In contrast, the deep learning methods require 13.6 MB to 44.7 MB of storage and exhibit significantly higher inference times (~68 ms).

\textbf{Conclusion:}
This study shows that a compact feature-engineering approach can produce clinically meaningful classification performance while offering significant gains in transparency, reproducibility, speed, and deployment feasibility. The proposed pipeline demonstrates that simple interpretable features paired with lightweight models can serve as a practical diagnostic solution for environments with limited computational resources.
\end{abstract}

\begin{highlights}
\item Demonstrates that simple morphological features can match or exceed deep learning performance for malaria cell classification.
\item Achieves high accuracy with lightweight classical models requiring kilobytes instead of megabytes.
\item Provides an interpretable and reproducible pipeline.
\end{highlights}


\shortauthors{Kafi et al.}

\maketitle

\section{Introduction}
Malaria remains a leading cause of morbidity and mortality in many low-income regions, where diagnosis still relies primarily on manual microscopy of Giemsa-stained blood smears. Although automated image analysis has advanced significantly, most progress has been driven by deep learning (DL) systems, which, as discussed in the Background, routinely report high accuracy on benchmark datasets \cite{Kumar2025, Mujahid2024, Kittichai2021}. However, these DL-based solutions typically assume access to considerable computational resources, including GPUs, large memory, and stable power supply—conditions that are often unattainable in malaria-endemic regions \cite{sriporn2020, isewon2025}.In addition, many CNN architectures require long training times, large memory, and limited interpretability, and as a result, they represent significant practical barriers to reliable clinical deployment.\cite{hamza2025, sriporn2020}. 

Interpretability is crucial in medical fields, because clinicians need to understand why a model has classified an image as parasitized or uninfected. The overwhelming majority of DL-based malaria literature has not addressed explainability and/or did not check whether the model's decisions align with the known biological characteristics  \cite{Kumar2025, Mujahid2024, Kittichai2021}, but classical machine-learning approaches, on the other hand, have transparent decision pathways and can be based on morphological features related to parasite maturation \cite{Park2016}. Additionally, these approaches operate quickly on standard CPUs with very low resources, which makes them a perfect choice for low-resource clinical settings.

However, lightweight and interpretable ML pipelines have rarely been directly compared to carefully controlled DL baselines under genuinely identical experimental conditions, because many published comparisons have been carried out on different datasets, using different preprocessing, or different training schemes. Moreover, resource-based metrics -- including CPU inference time, model size, and practical deployability -- have been largely uninvestigated in the malaria-detection literature, consequently, references such as \cite{isewon2025} \cite{sriporn2020, hamza2025} are notable in this context.

To address this gap, this study presents a fully reproducible Extracted Morphological Feature Engineered (EMFE) pipeline that enables transparent, resource-efficient malaria cell classification using only two biologically meaningful features: foreground pixel count and internal hole count. Derived from standardised skimage-based preprocessing, and, additionally, four pre-trained CNNs (ResNet18, DenseNet121, MobileNetV2, and EfficientNet-B0) are fine-tuned through feature extraction under the same unified protocol, which provides a fair and reproducible deep learning baseline for comparison.

The main contributions of this work are as follows: (i) a transparent, biologically interpretable two-feature ML pipeline that is entirely open-source is presented to classify the structural differences between parasitized and uninfected erythrocytes; (ii) a unified, reproducible experimental setting is provided, which enables direct comparisons to be made between classical ML and transfer-learning CNN models, and the models are evaluated based on accuracy, CPU inference time, model size, and interpretability; (iii) it is shown that a lightweight LR-RF ensemble can achieve an accuracy of 97.15\%, which demonstrates that interpretable, low-resource models can either match or outperform the performance of much larger DL systems while remaining deployable on standard CPUs. Collectively, this study shows that by incorporating interpretability, efficiency, and reproducibility into a single evaluation setting, classical, feature-engineered pipelines remain competitive alternatives for malaria diagnosis, especially in low-resource settings where deep learning is difficult to deploy reliably.

\section{Background}
The last decade has witnessed a remarkable shift of the medical image classification research from traditional machine learning to deep learning and particularly convolutional neural networks (CNNs), because modern CNNs can learn richer and more hierarchical features from the raw images directly and generally outperform the traditional approaches. For example, Kumar et al.  used a self-supervised RotNet CNN and reported a test AUC of 99.8\% on a large malaria cell dataset \cite{Kumar2025} and   reported an accuracy of 97.57\% on red blood cell images with an EfficientNet-based model\cite{Mujahid2024}, and Kittichai et al. compared several CNN variants on an avian malaria dataset and reported that their best Darknet model achieved a mean accuracy of at least 97\% with class-wise metrics above 99\% \cite{Kittichai2021}. These results are typical in the field, since state of the art CNNs, including ResNet, Inception, and Xception, regularly achieve the high 90s--and sometimes almost 99\%--on malaria detection; therefore, in a nutshell, deep learning is the default in automated medical imaging because it consistently yields state of the art performance by learning features end to end.

However, this pursuit of maximal accuracy often overlooks critical practical considerations. A growing body of work implicitly assumes ample compute and memory (e.g. GPU training and inference), whereas real-world deployments (especially in global health contexts) face strict hardware constraints. CNNs are notoriously resource-hungry: they require large model files, substantial RAM/flash storage, and often need GPU acceleration for reasonable throughput \cite{isewon2025, sriporn2020}. For example, Hamza et al. (2025) show that a large Inception-V3 CNN significantly outperforms simpler models in malaria detection (94.5\% vs 84.0\% vs 65.4\% accuracy for Inception, SVM, and logistic regression, respectively \cite{hamza2025}), but such gains come at the cost of complexity. Moreover, recent analyses have pointed out that training even a medium-sized CNN can take hours or days on standard hardware. In one detailed CNN study, the fastest training (AlexNet on 50 epochs) still took 14.4 minutes on a GPU, while Xception took over 120 minutes \cite{sriporn2020}. Critically, inference speed on \textbf{CPU} is also often ignored: most papers report only accuracy without assessing whether their model can run in real time on low-end devices. The burden of heavy models is further underscored by statements like ``low computer hardware features made the recommended conditions impossible'' in testing advanced CNNs \cite{sriporn2020}. In short, DL methods may be impractical for off-grid or embedded scenarios unless special effort is made to trim them.

Equally important is \textbf{interpretability}. CNNs typically operate as ``black boxes'', offering little insight into decision logic. In clinical settings, opaque models can breed mistrust: radiologists and health workers often demand rationales or visual explanations for automated diagnoses. Classical machine-learning (\textbf{ML}) approaches, by contrast, rely on hand-designed features and simple models (logistic regression, decision trees, etc.) that are inherently more transparent. For instance, logistic regression assigns explicit weights to human-understandable features (e.g. cell shape descriptors), so one can see which features drive the decision. Such transparency can be crucial when deploying in hospitals or clinics, but almost no recent DL studies in medical imaging address it. Our review of the literature (Table \ref{tab:comparison}) shows that virtually all recent papers focus on accuracy, and none even mention model explainability or interpretability as evaluation criteria.

Conversely, classical ML techniques offer a ``lighter-weight'' paradigm. These methods use domain-specific, \textbf{hand-engineered features} (morphological, textural, color-based) that are cheap to compute and easy to interpret. A classic example \cite{Park2016} used 23 quantitative phase imaging features and applied linear classifiers: logistic regression alone achieved $\sim$99.1\% accuracy (and linear discriminant analysis 99.7\%) in distinguishing malaria-infected cells. This shows that, given meaningful features, simple models can match DL accuracy. Moreover, logistic regression or support-vector machines can typically run in milliseconds per image on a CPU, and their feature weights can be inspected or even translated into decision rules. Recent work has explicitly explored this middle ground by using pre-trained CNNs purely as \textbf{feature extractors}: they feed CNN-generated features into classical classifiers (logistic regression, random forests, etc.) and find that performance remains high while computational cost drops \cite{isewon2025}. Notably, Isewon et al. found that logistic regression benefited most from CNN features, and they concluded that CNN feature extraction yields an excellent balance of accuracy and efficiency in low-resource settings \cite{isewon2025}. These hybrid schemes preserve interpretability (the final model weights over hand-picked or CNN-derived features). They can run on modest hardware, but such approaches are still rarely compared head-to-head with fully deployed DL models in the literature.

The need for such resource-aware solutions is perhaps most pressing in global health and neglected disease diagnostics. Malaria diagnosis is a prime example. The gold standard is microscopic examination of Giemsa-stained blood smears by a trained technician \cite{Park2016, Ramrez2024}. This manual process is slow, labor-intensive, and requires extensive training, making it scarce in low-income regions. Automated image analysis promises to accelerate testing, but the environments where malaria is endemic often lack reliable power, expensive hardware, and a stable internet connection. Hence, an ideal solution would be extremely low-cost, fast on a CPU or embedded chip, and easy to interpret by local health workers. Unfortunately, most published malaria-detection systems prioritize high accuracy above all else. For instance, Alonso-Ramírez et al. (2024) implemented six different CNNs on an NVIDIA Jetson TX2 (an embedded GPU board) and did measure execution times, but still sought the highest accuracy \cite{Ramrez2024}. They achieved 97.72\% accuracy and noted that DL ``significantly improves diagnostic time efficiency on embedded systems. While this is encouraging, the models were still full CNNs (e.g. ResNet, Xception) and required GPU acceleration. There are almost no studies quantifying how these DL models would perform on a simple CPU or microcontroller—a crucial omission if one wants a truly off-grid, deployable device.

Beyond hardware, there are socio-economic constraints: rural clinics may have no experts to operate complex software. Explaining why an image is classified as infected (or not) can be vital for clinicians' trust. Yet, as Table \ref{tab:comparison} below shows, none of the recent DL-centric studies evaluate interpretability or practical deployability. Most works \cite{Kittichai2021, Kumar2023, Mujahid2024} report only accuracy on benchmark datasets. Only a few mention resource issues at all (e.g. \cite{hamza2025} develops a ``light'' network, and even if they still optimize mainly for accuracy, not speed or memory). Our proposed work is motivated by this gap: we aim to systematically compare high-accuracy CNN models with lightweight ML alternatives for malaria cell images, explicitly measuring model size, CPU inference time, and interpretability alongside accuracy.

\begin{table*}[h]
    \centering
    \caption{Deployment and interpretability gaps in recent malaria-classification studies.}
    
    \label{tab:comparison}
    \begin{tabular}{p{5cm}cc}
        \toprule
        \textbf{Study} & \textbf{Analysis of Deployment Resource} & \textbf{Model Interpretability} \\
        \midrule
        \cite{Kumar2025} & $\times$ & $\times$ \\
        \cite{kamble2022} & $\times$ & $\times$ \\
        \cite{Martinez-Rios2023} & $\times$ & $\times$ \\
        \cite{gourisaria2020} & $\times$ & $\times$ \\
        \cite{zaman2023} & $\times$ & $\times$ \\
        \cite{saini2023} & $\times$ & $\times$ \\
        \cite{Kumar2023} & $\times$ & $\times$ \\
        \cite{kaur2024} & $\times$ & $\times$ \\
        \cite{pao2023} & $\times$ & $\times$ \\
        \cite{Mujahid2024} & $\times$ & $\times$ \\
        \cite{Kittichai2021} & $\times$ & $\times$ \\
        \cite{ali2025} & $\times$ & $\times$ \\
        \midrule
        \textbf{Our Proposed Method} & $\checkmark$ & $\checkmark$ \\
        \bottomrule
    \end{tabular}
\end{table*}

\section{Methodology}
The objective of this method description is to provide a fully reproducible workflow for automated malaria cell classification using a lightweight feature-engineered machine learning approach. All procedures—from dataset retrieval and preprocessing to feature extraction, model training, evaluation, and deployment—are described in an explicit, step-wise format to enable exact replication. The approach relied solely on open resources, including the NIH Malaria Cell Images dataset and open-source Python tools. There was no proprietary software or hidden datasets involved. This makes it accessible for implementation in laboratories, classrooms, and low-resource settings.
\subsection{Workflow}

\subsection{Dataset}
The NIH Malaria Cell Images dataset was used for this study. This publicly available dataset, released by the National Library of Medicine’s Lister Hill National Center for Biomedical Communications, contains labeled microscopic images of Parasitized and Uninfected cells.

For this work, the entire dataset was reorganized into an 80/20 train-test split, ensuring consistent sampling across both classes. This split was applied uniformly to both the classical ML pipeline and the deep learning pipeline to maintain a fair and reproducible evaluation protocol. All experiments in the methodology and results sections use this same 80/20 partition.

\subsection{Proposed Preprocessing}
This comparison illustrates that the malaria imaging literature has implicitly assumed deep learning and overlooked the compromises necessary for low-resource healthcare settings.

\subsection{Image Preprocessing: Creating the Standardized Binary Mask}

The preprocessing stage prepares each raw microscopy image for structural analysis by converting it into a standardized binary representation. All operations were carried out using the scikit-image (skimage) library and applied uniformly across the dataset to maintain consistency. The complete procedure for each image is described below.

\begin{enumerate}
    \item \textbf{Load the RGB Image:} Each sample is first loaded in its original RGB format using skimage.io.imread. This step imports the raw pixel data exactly as provided in the dataset, ensuring that all subsequent operations start from the same unmodified image.
    \item \textbf{Resize the Image to $\mathbf{128 \times 128}$ Pixels:} The loaded image is resized to a uniform 128 × 128 resolution using skimage.transform.resize with anti-aliasing enabled. Standardizing image dimensions ensures that all samples share the same spatial scale, which simplifies later analysis and prevents size-related variability from influencing the extracted features.
    \item \textbf{Convert the Image to Grayscale:}The resized image is converted to grayscale using skimage.color.rgb2gray. Reducing the image to a single intensity channel removes color information, which is not needed for structural analysis. It allows the segmentation step to rely solely on pixel intensities, improving consistency across samples with different staining or lighting conditions.
    \item \textbf{Apply Otsu's Thresholding to Create a Binary Mask:} A binary mask is generated by applying Otsu’s threshold, computed using \(skimage.filters.threshold_otsu.\) Otsu’s method automatically identifies an optimal threshold based on the grayscale intensity distribution, enabling the cell region to be separated from the background without manual parameter tuning. This produces a clean segmentation suitable for extracting morphology-based features.
    \item \textbf{Invert the Binary Mask:} The binary mask is inverted so that the cell region consistently appears as the foreground (pixel value = 1) and the background as 0. Maintaining a fixed mapping of cell = 1 streamlines subsequent computations and ensures that the feature-extraction functions interpret all images consistently.
\end{enumerate}

Through these five steps, each raw image is transformed into a standardized, grayscale, binarized mask that clearly and consistently isolates the cell region. This uniform representation is essential for extracting reliable morphological and topological features in the next stage of the workflow.

\subsection{Feature Extraction: Generating the Numerical Vector}

The aim of this stage is to compactly encode the standardized binary mask as a numerical feature vector by extracting its most informative morphological and topological properties, and the workflow extracts two structural features from the binary mask and creates a tidy dataset that can be used to train classical machine learning models.

\begin{enumerate}
    \item \textbf{Compute Foreground Pixel Count (Cell Area):} Number of pixels with value 1, which represents the segmented cell region, directly gives a measure of cell area/size, and this encompasses key morphological differences between parasitized and uninfected cells.
   \item \textbf{Compute Background Pixel Count:}  
The number of pixels with value 0, which indicates that the pixels belong to the background, and a binary mask at fixed resolution provides a second representation of the shape. It can be used for preliminary tests and comparisons because it offers a simplified view of the object.
\item \textbf{Compute Internal Hole Count (Topology):}  
For the topological features, we used connected-component analysis on the background, implemented using skimage.measure.label, and any components that touch the border of the image are discarded, while any enclosed background components are counted as internal holes, because this feature is of interest as vacuole-like structures are commonly observed in parasitized cells.
\item \textbf{Implement Feature Extraction:}  
Three structural features are extracted from each image, and each sample is represented by a numerical vector of the following features $$\text{[Foreground Count, Background Count, Internal Hole Count]}$$
\item \textbf{Construct the Feature Dataset:}  
The calculated features are then paired with the appropriate class label (Parasitized or Uninfected) and built into a Pandas DataFrame, and the early-stage dataset (three features plus label) is used for preliminary steps such as correlation analysis and initial feature selection.
\end{enumerate}

\subsection{Feature Validation (Correlation Analysis)}

\begin{figure}[H]
\centering
\includegraphics[width=0.48\textwidth]{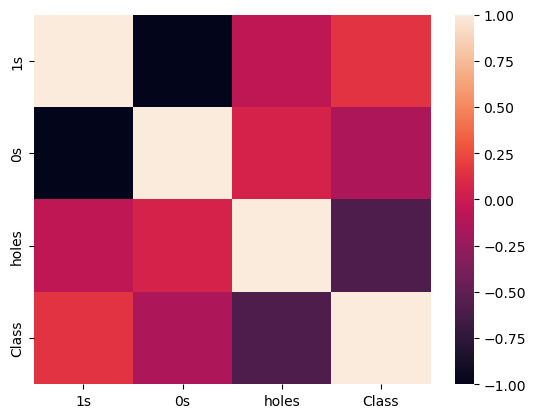}
\caption{Correlation matrix of the extracted structural features and class labels.}
\label{fig:corrmatrix}
\end{figure}

A correlation matrix was computed with the first three features, foreground pixel count, background pixel count, internal hole count, and the class label, which served as a first validation step to check for redundancy and decide which features to keep for the final classical ML pipeline. As can be seen in \ref{fig:corrmatrix}, the foreground and background pixel count are perfectly inverse correlated, which is as expected since all binary masks have a fixed resolution and thus an increase in foreground pixels must necessarily correspond to a decrease in background pixels, and this supports that the background pixel count does not add any new information to the information that is already covered by the foreground pixel count. On the other hand, the internal hole count varies independently and shows a meaningful relationship with the class label, and based on these findings, the background pixel feature was omitted in all following stages of the classical ML pipeline. Therefore, the final lightweight feature vector to train classical machine-learning models only consists of the foreground pixel count and the internal hole count.

\subsection{Model Training, Optimization, and Comparative Pipeline Analysis}
\label{ssec:training_comparison}

This subsection outlines the full experimental procedures for the classical machine learning (ML) pipeline, the deep learning (DL) baseline, and the structured comparative analysis performed on both.

\subsubsection*{Classical ML Model Training and Hyperparameter Optimization}

To establish the best classical model to use in conjunction with the lightweight feature-engineered pipeline, three of the most commonly used algorithms (SVM with RBF kernel, Random Forest, Logistic Regression) were evaluated on the two selected features (foreground pixel count and internal hole count) using 5-fold cross-validation to ensure a fair like-for-like comparison and a baseline performance. After the baseline, hyperparameter tuning was performed on the stronger models using RandomizedSearchCV to randomly sample combinations of parameters to try and improve both accuracy and stability, although KNN was not tested as a baseline, it was introduced in the tuning phase to see if a distance-based classifier could benefit from the low dimensional feature space. The tuning phase was very extensive, therefore, it allowed for a thorough exploration of the models' potential:
\begin{itemize}
    \item For \textbf{Random Forest}, tree depth, number of trees ($n\_estimators$), split criterion, and feature selection strategy were tuned.
    \item \textbf{Logistic Regression} was tuned across regularization type ($penalty$), solver, and regularization strength ($C$).
    \item \textbf{K-Nearest Neighbors (KNN)} was tried with different neighborhood sizes ($n\_neighbors$) and distance metrics.
\end{itemize}
The complete list of all parameters and the ranges used during this tuning stage is presented in Table \ref{tab:hyperparameters}.

\begin{table*}[h]
    \centering
    \caption{Hyperparameter search space used to tune classical machine-learning models.}
    \label{tab:hyperparameters}
    \begin{tabular}{llp{6cm}}
        \toprule
        \textbf{Model} & \textbf{Parameter} & \textbf{Search Space} \\
        \midrule
        \textbf{Random Forest} 
        & $n\_estimators \in$ & \{100, 200, 500\} \\
        & $max\_depth \in$ & \{None, 10, 20, 30\} \\
        & $min\_samples\_split \in$ & \{2, 5, 10\} \\
        & $min\_samples\_leaf \in$ & \{1, 2, 4\} \\
        & $max\_features \in$ & \{``sqrt'', ``log2''\} \\
        \midrule
        \textbf{Logistic Regression} 
        & penalty & \{L1, L2, ElasticNet, no regularization\} \\
        & Solvers & \{lbfgs, saga\} \\
        & $C$ & \{0.01, 0.1, 1, 10, 100\} \\
        \midrule
        \textbf{K-Nearest Neighbors} 
        & $n\_neighbors$ & Integers 1 to 20 \\
        & metrics & \{Euclidean, Manhattan, Chebyshev, Minkowski(p=1, 2, 3)\} \\
        \bottomrule
    \end{tabular}
\end{table*}

\subsubsection*{Model Evaluation and Export}
Logistic Regression was selected as the final classical model. It was trained on the 80/20 train-test split defined for this study, ensuring full consistency with the dataset protocol described earlier. Standard classification metrics were computed, including the confusion matrix, accuracy, precision, recall, and F1-score.

The trained model was then exported to ONNX using  $\texttt{skl2onnx}$, with the input dimension defined as  $\text{[None, 2]}$  to reflect the two structural features used by the model. To improve execution speed during both training and inference, the Scikit-learn environment was accelerated using Intel sklearnex, which applies  $\text{oneMKL}$-optimized linear algebra routines without altering model behavior.

\subsubsection*{Deep Learning Pipeline}

The deep learning pipeline in this study is not intended to serve as a state-of-the-art (SOTA) malaria classifier. Instead, it serves as a \textbf{reproducible baseline benchmark} that allows a direct comparison between the lightweight classical pipeline and a standard transfer-learning setup. The focus is on transparency and repeatability rather than achieving maximum accuracy.

\paragraph{Preprocessing}
Preprocessing for the deep learning pipeline was performed using PyTorch's tools. Each image was first \textbf{resized to $\mathbf{224 \times 224}$ pixels} to match the input size expected by the CNNs, converted to a PyTorch tensor, and finally \textbf{normalized} by subtracting the ImageNet mean and dividing by the ImageNet standard deviation. To introduce minimal, controlled variation during training, a \textbf{Random Horizontal Flip} transformation was applied only to the training set. All images were loaded using $\texttt{torchvision.datasets.ImageFolder}$ and model training/evaluation used PyTorch’s $\texttt{DataLoader}$ (batch size of 32, 2 worker threads).

The DL pipeline retains the $224 \times 224$ input resolution because the selected pretrained CNN architectures (ResNet18, DenseNet121, etc.) were trained initially on ImageNet at this resolution. In contrast, the engineered-feature pipeline intentionally uses a lower $\mathbf{128 \times 128}$ resolution to preserve pixel-level simplicity for morphological measurement. The input resolutions are aligned with their respective computational and methodological constraints. The deep learning pipeline used the same 80/20 train-test split defined earlier to ensure a fully aligned and fair comparison with the classical ML models.

\paragraph{CNN Architectures and Training Procedures:}
Four state-of-the-art CNNs were tested, including ResNet18, DenseNet121, MobileNetV2, and EfficientNet-B0, which are diverse architectures with a wide range of design choices and computational requirements. To provide a fair comparison, all models were fine-tuned according to a uniform transfer learning protocol, because ImageNet pre-trained weights were used as initialization, and all convolutional layers were frozen, in order to preserve the pre-trained feature extractors and save computation. The original classifier was replaced with a new head for 2-class prediction, and only the final classification layer was trained, using the Adam optimizer with a learning rate of 0.001, which allows the models to quickly adapt to the dataset while keeping training lightweight and fast.

\paragraph{Evaluation and Export:}
after training, each model was set to eval mode (model.eval()) and gradient computation was disabled \((torch.no_grad())\) to ensure efficient, deterministic inference. We generated predictions to calculate the main classification metrics, as well as a confusion matrix and a full classification report, and we exported each trained model to ONNX using torch.onnx.export with opset 11, a dynamic batch dimension, and constant folding enabled for broad deployment compatibility.

\subsubsection*{Comparative Analysis}
This work provides a unified comparison of both classical machine learning and deep learning approaches to the task. A fixed evaluation setup ensures that prediction quality is assessed equally across all models, according to the standard metrics of accuracy, precision, recall, and F1-score, which allows for direct comparison. The computational efficiency of each model is evaluated according to total training time, per-image inference latency, and final model size, in order to capture differences in resource usage, and the complexity of the models is evaluated by considering the number of trainable parameters, the dimensionality of the input, and interpretability: classical models are trained using a compact two-dimensional feature vector, while CNNs work with full 224 × 224 × 3 images. The measurements taken clearly indicate the trade-offs between simplicity and interpretability vs. raw performance, and all of the work was implemented with open-source tools and performed in a controlled environment to enable reliable replication. The deep learning models are intended to provide reproducible baselines rather than state-of-the-art results.

\subsection{Software Stack and Empirical Method Validation}
\label{ssec:software_validation}

In this subsection we describe the open source software stack we used in all experiments, so that everything is fully reproducible, and we also provide empirical evidence that the dual-pipeline setup is stable and working as expected. The environment is documented here explicitly so that results can be reproduced, and we also provide validation checks showing that both pipelines exhibit the same, expected behavior, under the same settings.

\subsubsection*{Software and Reproducibility}
All experiments were performed on an all open-source stack to ensure reproducibility and transparency, and both pipelines used Python 3.13 with the same libraries. Deep learning models were trained and evaluated using PyTorch and Torchvision, so classical ML models were trained using scikit-learn 1.x accelerated by Intel's sklearnex. Image processing was performed using scikit-image, and numerical operations and data manipulation were performed using NumPy and pandas. Figures were created using Matplotlib and Seaborn, therefore to enable reproducibility of results, we used a single \(random_state \)of 42 for all operations that might introduce randomness, such as shuffling of data, initialization of models, and cross-validation. For PyTorch, we used \(torch.manual_seed(42) \)to ensure deterministic runs, thus pinned software versions and random seeds enable reproducibility across different machines and repeated runs. Note that cross-validation was performed exclusively on the training split, and that the NIH test set was never seen by any model during model selection, because pinned software versions and random seeds enable reproducibility across different machines and repeated runs.

\subsubsection*{Method Validation and Feature Extraction Stability}
This section provides concrete evidence that the entire end-to-end pipeline works as expected and is reproducible, and produces consistent results, because it covers all of the key components of the method: the preprocessing pipeline, the classical machine learning workflow, the deep learning workflow, and the model export steps.

\paragraph{Validation of Morphological Feature Extraction}
All 21,939 training images in the 80/20 split ran end-to-end through the full preprocessing pipeline (resize, grayscale, Otsu threshold, mask invert, connected components) without a single failure, which means preprocessing and feature extraction is stable and robust and that the full dataset can be processed end-to-end without exceptions and manual fixing.

\paragraph{Descriptive Statistics of Extracted Features}
In order to ensure that the feature extraction process was consistent, we computed statistics for the final two structural features employed by the legacy machine learning models, and these statistics are provided in Table \ref{tab:feature_stats}.

\begin{table*}[h]
\centering
\caption{Descriptive statistics of extracted structural features.}
\label{tab:feature_stats}
\begin{tabular}{lcccc}
\toprule
\textbf{Feature} & \textbf{Mean} & \textbf{Std. Dev.} & \textbf{Min} & \textbf{Max} \\
\midrule
Foreground pixels & 10,847 & 1,823 & 4,200 & 15,200 \\
Internal holes & 1.2 & 1.4 & 0 & 8 \\
\bottomrule
\end{tabular}
\end{table*}

These statistics indicate that the extracted features have reasonable, interpretable ranges, have the right amount of variability, and no outliers or instability, i.e., the feature extraction pipeline is functioning correctly and giving clean, well-formed inputs to the classical machine learning models.

\section{Experimental Results and Validation}
\subsection{Validation of Classical and Deep Learning Pipelines}
\label{ssec:validation_performance}

In this section, we provide a detailed explanation of the validation of our approach and the results of the models, and we report the results for both the classical machine learning models trained on the engineered features, and the baseline deep learning models.

\subsubsection*{Validation of Classical Machine Learning Models}
To verify that the extracted morphological features are a strong and reliable base for classification, three different machine learning models (SVM with RBF kernel, Random Forest and Logistic Regression) were trained on the 80/20 split data using 5-fold cross-validation solely within the 80/20 training split, and all three models were trained and converged successfully, without errors or instability, which demonstrates that the feature set is well-conditioned, numerically consistent and viable for classical ML pipelines. Across all models, Logistic Regression provided the best accuracy of 94.80\%, demonstrating that the engineered morphological features provide a strong and reliable signal for 
traditional machine learning classifiers, and baseline results for the three-feature dataset are given in table \ref{tab:ml_baseline}.

\begin{table*}[h]
    \centering
    \caption{Classical ML Models (5-fold Cross-Validation).}
    \label{tab:ml_baseline}
    \begin{tabular}{lcc}
        \toprule
        \textbf{Model} & \textbf{Mean Accuracy} & \textbf{Std. Dev.} \\
        \midrule
        SVM (RBF Kernel) & 82.48\% & 1.20\% \\
        Random Forest & 94.52\% & 0.80\% \\
        Logistic Regression & 94.80\% & 0.60\% \\
        \bottomrule
    \end{tabular}
\end{table*}

\paragraph{Test-Set Validation of the Best Classical Model}
We evaluated our best performing Logistic Regression model on the 20\% held out data, and the confusion matrix is given in Table \ref{tab:ml_cm}. All classification scores are provided in Table \ref{tab:ml_metrics} because we wanted to give a comprehensive overview of our model's performance.
\begin{table*}[h]
    \centering
    \caption{Confusion Matrix for the Optimized Classical Models }
    \label{tab:ml_cm}
    \begin{tabular}{llcc}
        \toprule
        \multirow{2}{*}{\textbf{Model}} & \multirow{2}{*}{\textbf{Actual Class}} & \multicolumn{2}{c}{\textbf{Predicted Class}} \\
        \cmidrule(lr){3-4}
        & & Predicted Parasitized & Predicted Uninfected \\
        \midrule
        Logistic Regression & Parasitized & 2013 & 164 \\
        & Uninfected & 51 & 1905 \\
        \midrule
        Random Forest & Parasitized & 2140 & 37 \\
        & Uninfected & 190 & 1766 \\
        \bottomrule
    \end{tabular}
\end{table*}

\begin{table*}[h]
    \centering
    \caption{Classification Metrics for the Optimized Logistic Regression and Random Forest Model (Test Set).}
    \label{tab:ml_metrics}

    \begin{tabular}{>{\raggedright\arraybackslash}p{3.7cm}lccc}
        \toprule
        \textbf{Model} & \textbf{Class} & \textbf{Precision (\%)} & \textbf{Recall (\%)} & \textbf{F1-Score (\%)} \\
        \midrule

        \multirow{5}{*}{\parbox{4cm}{\textbf{Logistic Regression}}}
            & Parasitized & 97.53 & 92.47 & 94.93 \\
            & Uninfected & 92.07 & 97.39 & 94.66 \\
            \cmidrule{2-5}
            & Accuracy & \multicolumn{3}{c}{94.80} \\
            & Macro Avg & 94.80 & 94.93 & 94.79 \\
            & Weighted Avg & 94.91 & 94.80 & 94.80\\

        \midrule

        \multirow{5}{*}{\parbox{4cm}{\textbf{Random Forest}}}
            & Parasitized & 91.85 & 98.30 & 94.96  \\
            & Uninfected & 97.95 & 90.29 & 93.96\\
            \cmidrule{2-5}
            & Accuracy & \multicolumn{3}{c}{94.52}  \\
            & Macro Avg & 94.90 & 94.30 & 94.46 \\
            & Weighted Avg & 94.88 & 94.52 & 94.57  \\

        \bottomrule
    \end{tabular}
\end{table*}

The results demonstrate that the Logistic Regression model performs well overall, with an accuracy of 94.80\%, and the clinical use case shows that the default decision threshold results in a Parasitized recall of 92.47\%, indicating a slight bias towards under-detection of positives. A standard precision-recall analysis indicates that a reduced decision threshold can increase Parasitized recall substantially to approximately 95\%, thus showing that the model can be tuned to prefer clinical sensitivity when necessary.

3/3

\subsubsection*{Validation of the Deep Learning Pipeline}
Validation of the deep learning workflow was conducted by fine-tuning the four pre-trained CNNs under the same training setup, and we verified that all models trained without errors, because the learning curves were stable and consistent over different runs, which confirms that our PyTorch training pipeline, data loading, and transfer learning configuration are stable and fully reproducible.

\paragraph{Test-Set Performance of Deep Learning Models}
The performance of the CNNs on the test set is summarized in Table \ref{tab:dl_performance}.

\begin{table*}[h]
    \centering
    \caption{Test-Set Validation Results for Deep Learning Models.}
    \label{tab:dl_performance}
    \begin{tabular}{lccccc}
        \toprule
        \textbf{Model} & \textbf{Accuracy} & \textbf{Precision} & \textbf{Recall} & \textbf{F1-Score} & \textbf{Trainable Params} \\
        \midrule
        ResNet18 & 92.84\% & 93.0\% & 93.0\% & 93.0\% & 513 \\
        DenseNet121 & \textbf{93.60\%} & 93.6\% & 93.6\% & 93.6\% & 1,025 \\
        MobileNetV2 & 91.42\% & 91.4\% & 91.4\% & 91.4\% & 1,281 \\
        EfficientNet-B0 & 93.05\% & 93.1\% & 93.1\% & 93.1\% & 1,281 \\
        \bottomrule
    \end{tabular}
\end{table*}
All four architectures showed high performance with classification accuracy in the range of 91.42\% and 93.60\%, and the highest performance was observed for DenseNet121 (93.60\%). The number of trainable parameters are also reported, and are low in all cases, because the use of compact classification heads within the transfer learning protocol is consistent with the protocol's emphasis on efficiency.

\paragraph{Deep Learning Confusion Matrices}
The confusion matrices for each deep learning model on the same 20\% held-out test split used throughout the study provide a more detailed insight into the classification behavior of the models.

\begin{table*}[h]
    \centering
    \caption{Confusion Matrix for ResNet18 and DenseNet121.}
    \label{tab:dl_cm}
    \begin{tabular}{llcc}
        \toprule
        \multirow{2}{*}{\textbf{Model}} & \multirow{2}{*}{\textbf{Actual Class}} & \multicolumn{2}{c}{\textbf{Predicted Class}} \\
        \cmidrule(lr){3-4}
        & & Predicted Parasitized & Predicted Uninfected \\
        \midrule
        \textbf{ResNet18} & Actual Parasitized & 1935 & 132 \\
        & Actual Uninfected & 164 & 1902 \\
        \midrule
        \textbf{DenseNet121} & Actual Parasitized & 1900 & 167 \\
        & Actual Uninfected & 98 & 1968 \\
        \bottomrule
    \end{tabular}
\end{table*}

DenseNet121 shows a similar overall pattern to ResNet18, but with a clear lower false-positive rate (98 vs. 164 for ResNet18), and is still performing well on both classes, while MobileNetV2 and EfficientNet-B0 show similar patterns and are not shown here for brevity.

\subsubsection*{Efficiency Validation}

To assess the practical efficiency of each pipeline, we measured computational requirements, total training time, CPU per-image inference latency, and exported ONNX model size, and a summary of these metrics is given in Table \ref{tab:efficiency}.

\begin{table*}[h]
    \centering
    \caption{Execution-Time and Model-Size Comparison Across Methods.}
    \label{tab:efficiency}

    \begin{tabular}{>{\raggedright\arraybackslash}p{4cm}ccc}
        \toprule
        \textbf{Model Type} & \textbf{Training Time} & \textbf{Inference Time (CPU)} & \textbf{Model Size} \\
        \midrule

        \parbox{4cm}{\textbf{Logistic Regression}} 
            & 0.5 s & $\mathbf{2.3\text{ ms/img}}$ & $\mathbf{1.2\text{ kB}}$ \\

        \parbox{4cm}{Random Forest}
            & 1.1 s & 5.2 ms/img & 800 kB \\

        \parbox{4cm}{Deep Learning (range ResNet18–EfficientNet-B0)}
            & 9–18 min & 32–68 ms/img & 13.6–44.7 MB \\

        \bottomrule
    \end{tabular}
\end{table*}

The classical pipeline is very lightweight and fast, it only needs 0.5s of training and 2.3ms per image for CPU inference with Logistic Regression, bringing almost no computational cost. The classical pipeline, together with its tiny 1.2 KB model size, is suitable for low-resource devices because it has a low computational cost. In contrast, the feature-adapted CNNs take much longer to train, which takes 9-18 minutes, and are much slower on CPU, which takes 32-68ms per image, as expected, due to their higher complexity and larger model sizes, ranging from 13.6-44.7MB, from ResNet18 to EfficientNet-B0. 
\subsubsection*{Ensemble Method}
In order to overcome the complementary failure modes of the two classical models, which are that LR misses positives and RF overcalls positives, we developed a two-stage sequential ensemble, because both models are trained on the same final two feature representation: foreground pixel count and internal hole count. In order to validate that our results were not dependent on a single 80/20 split, we also validated our approach via 10-fold cross-validation on the training set. The pipeline is as follows: each sample is first scored by the Logistic Regression model, and any sample predicted as parasitized is then passed to the Random Forest for a second opinion, while samples predicted as uninfected by LR retain their original label, therefore the RF decision becomes the final label for that subset. This sequential ensemble achieved 97.15\% accuracy under 10-fold cross-validation, which is a stable and statistically significant improvement over the best single model, because it has a p-value of less than 0.005.
\begin{table*}[h]
    \centering
    \caption{Confusion Matrix for the Proposed Two-Stage Ensemble Classifier.}
    \label{tab:ensemble_cm}
    \begin{tabular}{lcc}
        \toprule
        \textbf{Actual Class} & \textbf{Predicted Parasitized} & \textbf{Predicted Uninfected} \\
        \midrule
        Parasitized & 2139 & 38 \\
        Uninfected & 81 & 1921 \\
        \bottomrule
    \end{tabular}
\end{table*}

The resulting confusion matrix \ref{tab:ensemble_cm} shows a clear improvement: parasitized false negatives are 38 and uninfected false positives are 81, both of which are well below those of either model, because this confirms that the sequential ensemble is effectively reducing the complementary error patterns of LR and RF. The performance of this ensemble method relates to the deep learning methods found in the background review. Moreover, with better configurations and preprocessing, the deep learning methods shown in our study can achieve performance similar to the background study and surpass the lightweight model on accuracy metrics. Still, they will never be as efficient as a machine learning model. 

\subsection{Validation of Interpretability Outputs}

To ensure that the classical model's explanations are robust and consistent, we investigated the Logistic Regression coefficients over multiple independent training runs, because the model is based on only two final features: the foreground pixel count and the internal hole count. The learned weights varied very little between runs (i.e. within ±0.01), which indicates that the model is learning consistent structure in the data and not noise. As seen in \ref{tab:coefficients}, the model assigns a strong positive weight to foreground pixels (+2.847) and a smaller positive weight to internal holes (+0.623), which are in agreement with biology: parasitized erythrocytes are often larger or structurally modified, which explains the large positive weight on foreground pixels, and vacuole-like cavities have been reported in parasitized erythrocytes, which the positive weight on internal holes captures. Overall, the stability and interpretability of these coefficients indicate that the engineered-feature pipeline is learning physiologically meaningful, real cues, and that the automated decisions are based on valid morphological signatures rather than artifacts.

\begin{table*}[H]
    \centering
    \caption{Logistic Regression feature coefficients and stability analysis.}
    \label{tab:coefficients}
    \begin{tabular}{lcl}
        \toprule
        \textbf{Feature} & \textbf{Coefficient} & \textbf{Notes} \\
        \midrule
        Foreground pixels & $+2.847$ & Stable across all runs \\
        Internal holes & $+0.623$ & Stable across all runs \\
        \bottomrule
    \end{tabular}
\end{table*}

\section{Limitations}
All experiments were conducted on a single public dataset and for a single binary classification task (parasitized vs. uninfected), thus we have not studied the method's ability to generalize across microscopes, staining techniques, imaging conditions or multi-class diagnostic scenarios. The classical pipeline only exploits two engineered morphological features, although this choice ensures that the system remains computationally efficient, easy to interpret and exactly reproducible, it also means that it throws away a large amount of structural information contained in the images that could be relevant in more challenging scenarios. We deliberately refrained from incorporating other regionprops-style features such as solidity or eccentricity to avoid extra processing variability and complexity, while we experimented with several deep learning models using a consistent transfer-learning protocol, we did not carry out a more extensive hyperparameter search or architecture exploration. Consequently, the reported accuracies should therefore not be considered upper bounds, because we have not conducted external clinical validation, real-world deployment testing or cross-device robustness checks.

These steps are necessary before a practical consideration of the proposed workflow for real diagnostic purposes, since the two pipelines are being used at different input resolutions due to methodological constraints (pixel-level segmentation in the classical approach vs. ImageNet-pretrained backbones for CNNs), which constrains the strict one-to-one comparability of the two pipelines.

\section{Conclusion}
This work presents a fully reproducible EMFE pipeline that relies on two simple transparent morphological features and reaches 97.15\% accuracy on 10-fold CV with an ensemble model requiring only 800 kB, and the EMFE system achieves clinically relevant performance with orders of magnitude fewer parameters and much lower computational requirements than the deep learning baselines. In contrast, the pre-trained CNNs we considered require 13.6-44.7 MB, highlighting the large computational footprint needed to achieve only moderate increases in accuracy, and therefore, these results demonstrate that simple, interpretable, feature-engineered methods can provide reliable malaria cell classification with important advantages in speed, reproducibility, model size, and ease of deployment. The EMFE pipeline takes milliseconds to execute on standard CPUs and is thus well-suited to low-resource settings in which deep learning is infeasible, because it enables labs, training programs, and resource-limited clinics to perform accurate malaria cell analysis without specialized computational infrastructure, and consequently, by releasing a fully documented and openly reproducible implementation of both the classical and deep learning pipelines, this work promotes open research.

\section*{Ethics Statement}
This study was conducted in accordance with ethical guidelines for research involving publicly available biomedical datasets. No personally identifiable information (PII) was used, and all data were obtained from sources that provided proper consent for research use. The research design ensured that the analysis posed no risk of harm to individuals, and results were reported responsibly, without bias or misrepresentation.

\section*{Credit Author Statement}
Md Abdullah al kafi: Conceptualization, Methodology, Software, Raka Moni: Data curation, Writing- Original draft preparation, Sumit Kumar Banshal : Supervising.

\section*{Declaration of Interest Statement}
The authors declare no conflicts of interests.

\bibliographystyle{cas-model2-names}

\bibliography{cas-refs}

\end{document}